\documentclass[10pt,twocolumn,letterpaper]{article}

\usepackage{cvpr}              

\usepackage{graphicx}
\usepackage{amsmath}
\usepackage{amssymb}
\usepackage{booktabs}
\usepackage{listings}
\usepackage{color}
\usepackage{multirow}
\definecolor{codegreen}{rgb}{0,0.5,0}
\definecolor{codeblue}{rgb}{0,0,0.9}
\definecolor{codeblues}{rgb}{0,0,0.4}
\definecolor{codegray2}{rgb}{0.4,0.4,0.4}
\definecolor{codegray}{rgb}{0.9,0.9,0.9}
\definecolor{codepurple}{rgb}{0.58,0,0.82}
\definecolor{backcolour}{rgb}{0.95,0.95,0.92}
\definecolor{backcolour2}{rgb}{0.9,0.9,0.9}
\definecolor{codered}{rgb}{0.5,0,0}
\definecolor{textcodered}{rgb}{0.05,0.05,0.05}
\definecolor{palegray}{rgb}{0.98,0.98,0.99}

\definecolor{drawercolor}{RGB}{244,157,78}

\definecolor{reachcolor}{RGB}{153, 51, 102}

\definecolor{staplercolor}{RGB}{80, 200, 180}

\usepackage{colortbl}  

\newcommand*{\imgintext}[1]{%
  \raisebox{-.3\baselineskip}{%
    \centering\includegraphics[
      height=\baselineskip,
      width=\baselineskip,
      keepaspectratio,
    ]{#1}%
  }%
}

\usepackage[pagebackref,breaklinks,colorlinks]{hyperref}
\newcommand{\model}{EC$^{2}$\xspace}
\usepackage[capitalize]{cleveref}
\crefname{section}{Sec.}{Secs.}
\Crefname{section}{Section}{Sections}
\Crefname{table}{Table}{Tables}
\crefname{table}{Tab.}{Tabs.}

\def\cvprPaperID{3161} 
\def\confName{CVPR}
\def\confYear{2023}

\begin{document}

\title{EC$^2:$ Emergent Communication for Embodied Control
}
\author{
Yao Mu\textsuperscript{1},
Shunyu Yao \textsuperscript{2},
Mingyu Ding \textsuperscript{1},
Ping Luo\textsuperscript{1},
Chuang Gan\textsuperscript{3,4}\\
\textsuperscript{1}The University of Hong Kong \
\textsuperscript{2} Princeton University \
\textsuperscript{3}UMass Amherst, \textsuperscript{4}MIT-IBM Watson AI Lab\\
{ymu, myding, pluo}@cs.hku.hk \
shunyuy@princeton.edu\
ganchuang@csail.mit.edu
}

\maketitle


\begin{abstract}
Embodied control requires agents to leverage multi-modal pre-training to quickly learn how to act in new environments, where video demonstrations contain visual and motion details needed for low-level perception and control, and language instructions support generalization with abstract, symbolic structures. While recent approaches apply contrastive learning to force alignment between the two modalities, we hypothesize better modeling their complementary differences can lead to more holistic representations for downstream adaption.
To this end, we propose \textbf{E}mergent \textbf{C}ommunication for \textbf{E}mbodied \textbf{C}ontrol (\model), a novel scheme to pre-train video-language representations for few-shot embodied control. The key idea is to learn an unsupervised ``language'' of videos via emergent communication, which bridges the semantics of video details and structures of natural language. We learn embodied representations of video trajectories, emergent language, and natural language using a language model, which is then used to finetune a lightweight policy network for downstream control.
Through extensive experiments in Metaworld and Franka Kitchen embodied benchmarks, \textbf{\model} is shown to consistently outperform previous contrastive learning methods for both videos and texts as task inputs.
Further ablations confirm the importance of the emergent language, which is beneficial for both video and language learning, and significantly superior to using pre-trained video captions.
We also present a quantitative and qualitative analysis of the emergent language and discuss future directions toward better understanding and leveraging emergent communication in embodied tasks.

\end{abstract}

\section{Introduction}
\label{sec:intro}


We study the problem of few-shot embodied control, where an embodied agent needs to execute language instructions or follow video demonstrations given only a few examples in a new environment. Such a capability of fast adaption is key to practical robotic deployment, as it is non-scalable and expensive to collect extensive action-labeled trajectories for each new application scenario. Instead, agents need to leverage pre-trained video and language representations to quickly learn how to act, and generalize across high-level concepts (e.g.\,object and verb types) as well as low-level visual features.

\begin{figure}
     \centering
     \begin{subfigure}[b]{0.45\textwidth}
         \centering
         \includegraphics[width=\textwidth]{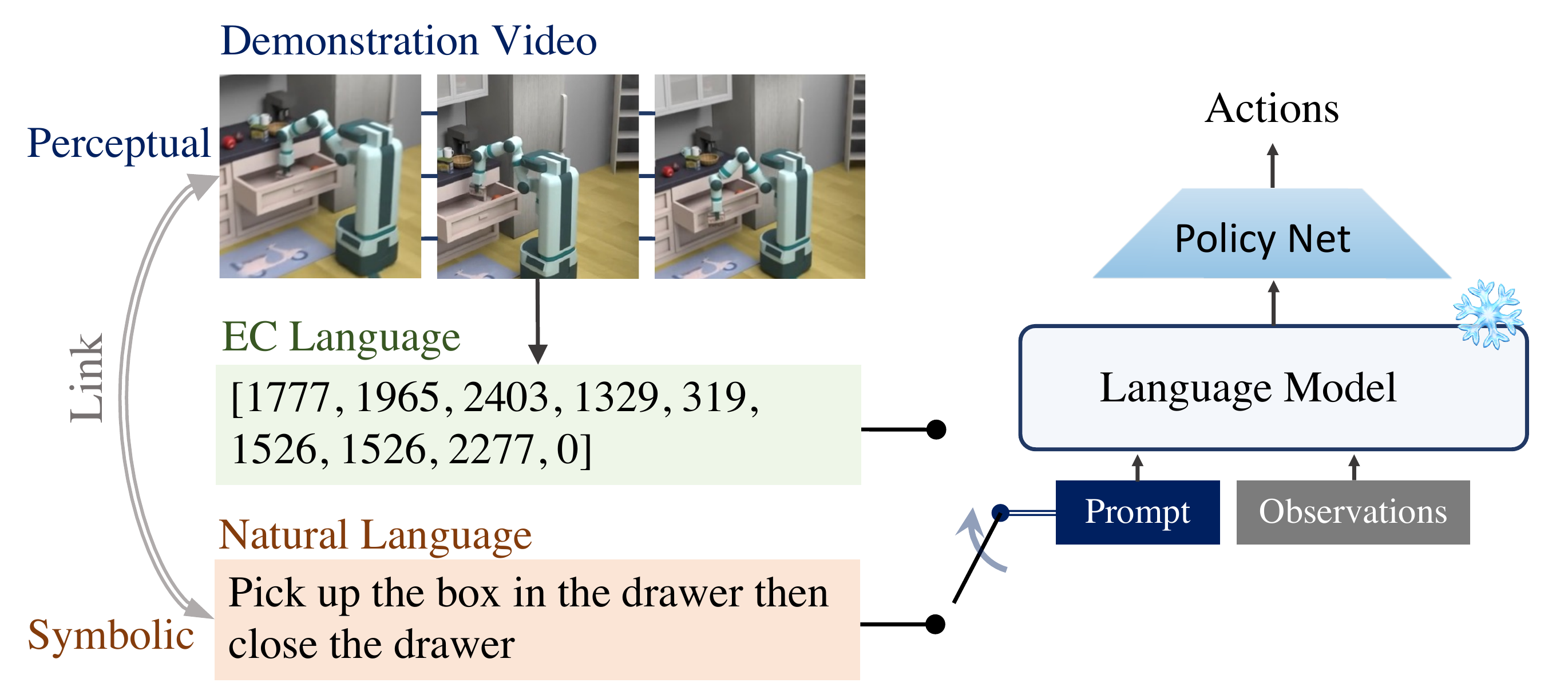}
         \caption{Emergent language for embodied control.}
         \label{fig:key idea}
     \end{subfigure}
     \hfill
     \begin{subfigure}[b]{0.45\textwidth}
         \centering
         \includegraphics[width=\textwidth]{ 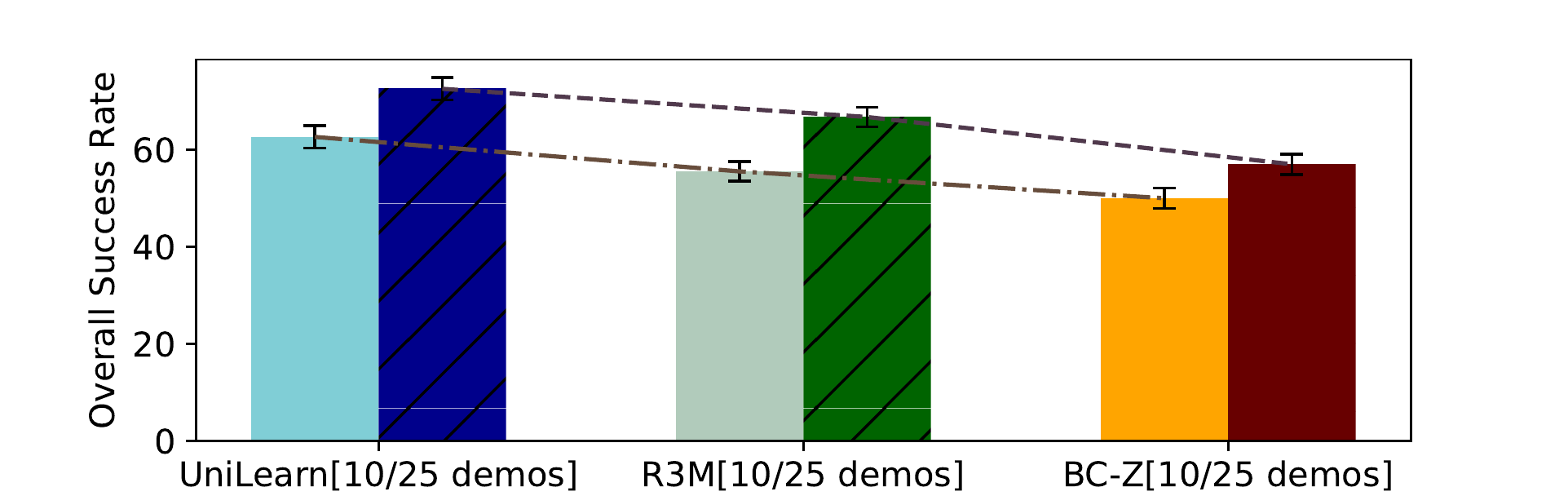}
         \caption{Performance comparison on overall success rate across all tasks.}
         \label{fig:Overall Success rate}
     \end{subfigure}
     \vspace{-5pt}
        \caption{Overview of \model. The key idea of \model is to build the link between perceptual grounding and symbolic concept via emergent communication (EC) language. We learn to extract embodied representation via language model utilizing emergent language or natural language as prompt and visual observation as input for downstream embodied control tasks. The actions are generated by a lightweight policy network containing a few MLP~\cite{ramchoun2016multilayer} layers that map the learned frozen embodied representation into action space. Extensive experiments show that \model outperforms existing methods in Metaworld~\cite{yu2020meta}  and Franka Kitchen~\cite{gupta2019relay} benchmarks.
}
\vspace{-10pt}
        \label{fig:first graph}
\end{figure}

How can we pre-train multi-modal representations of videos and texts for downstream embodied control in different domains? 
Inspired by the success of image-text models such as CLIP~\cite{radford2021learning}, recent work has investigated contrastive representation learning approaches using paired video-language data, with application toward robotic manipulation and control tasks~\cite{jang2022bc,nair2022r3m}.
However, simply aligning video and language pairs in the embedding space omits the difference between these two modalities: while videos contain more visual and motion details, language abstracts away key structures underlying the task.

For example, a video of opening a door contains details of approaching the handle, pressing the handle, pulling the door open, and so on, while the corresponding language description could be as simple as ``open a door'' ---  more detailed language descriptions such as ``press down the handle'' is usually not available in existing datasets and very expensive to manually label. 
But for downstream control purposes, a video and its abstract language description could present complementary benefits --- while the former is expressive in details needed for low-level control, the latter provides the structure for generalization across domains, tasks, and skills.
Leveraging such modality differences presents opportunities to better incorporate language and videos for embodied representation learning, marrying their respective benefits rather than forcing them to be aligned.

\begin{figure*}
     \centering
         \includegraphics[width=0.85\textwidth]{ 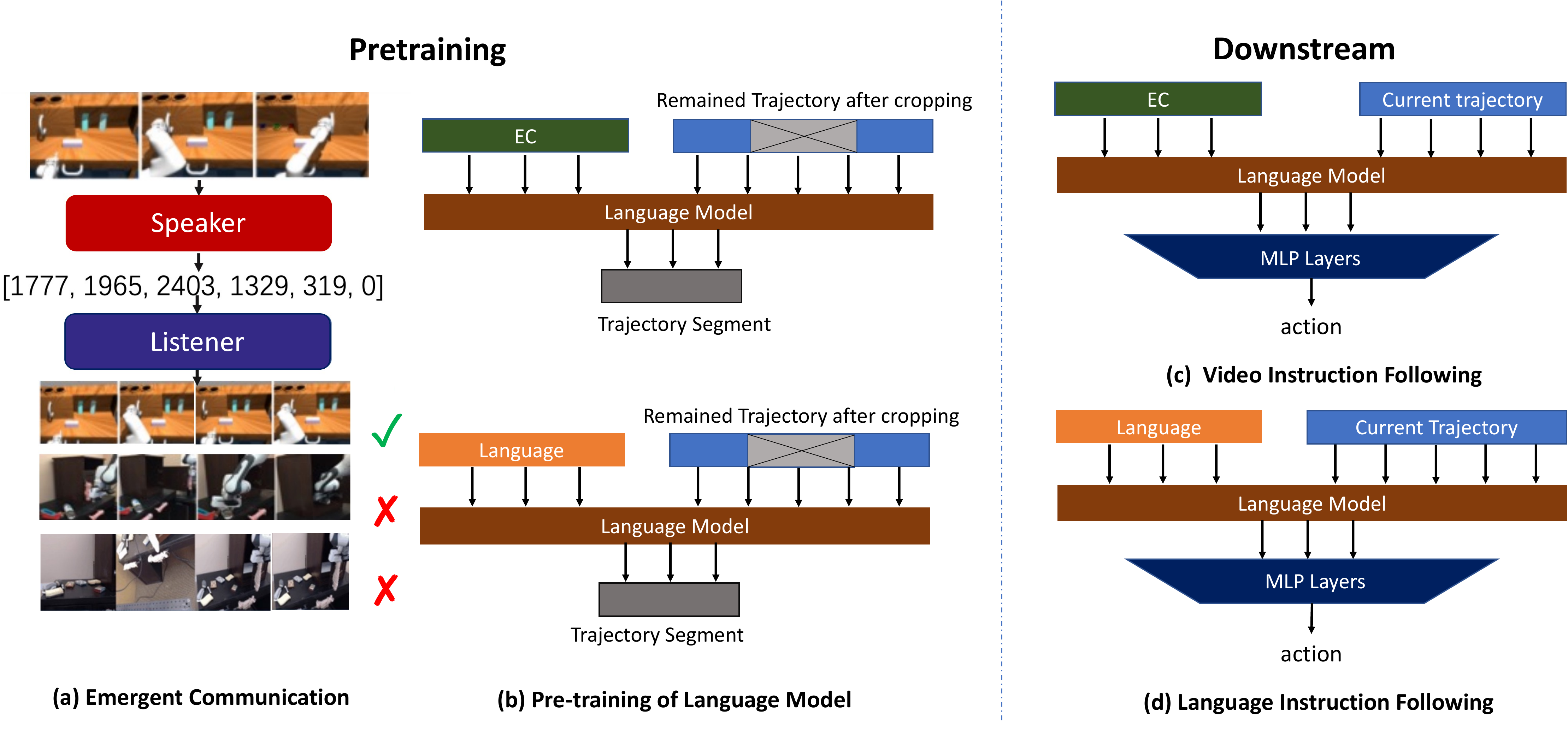}

    \vspace{-5pt}
        \caption{The overall framework of \model. \textbf{Pre-training:} \textbf{(a)} We first pre-train a speaker (\imgintext{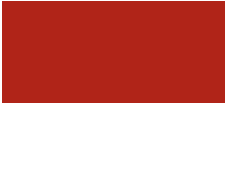}) and listener (\imgintext{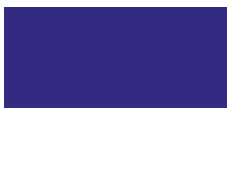}) via emergent communication game for emergent language generation and a language model (\imgintext{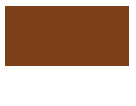}) to extract embodied representations. 
        \textbf{(b)} We generate the corresponding emergent language (\imgintext{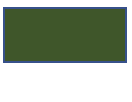}) for each demonstration video in the multi-modal dataset. Predicting the cropped part of a latent trajectory is used as an auxiliary task to help pre-training the language model. 
        The language model takes emergent language or natural language as prompt and the cropped trajectory (\imgintext{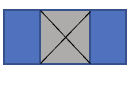}) as input and aims to predict the cropped part of the trajectory (\imgintext{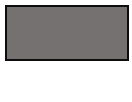}). The loss for the prediction using emergent language as prompt and natural language as prompt is calculated separately. 
        \textbf{Few-shot downstream policy learning:}
        In downstream imitation learning tasks, both the EC speaker and the pre-trained language model(\imgintext{icons/lang_m}) are frozen. The language model extracts effective embodied features with current trajectory (\imgintext{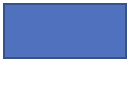} contains current state and historical information) as input. \textbf{(c)} For tasks with language instruction, we use natural language as the prompt, and \textbf{(d)} for tasks with video instruction, we use emergent language as the prompt.
        The output features of the language model are mapped into actions to mimic expert behavior by a few simple MLP~\cite{ramchoun2016multilayer} layers (\imgintext{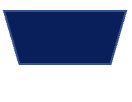}). }
        \vspace{-10pt}
        \label{fig:UniLearn frame}
\end{figure*}

One perspective to view the video-language difference is through the emergence of language~\cite{Nowak1999TheEO, Wagner2003ProgressIT}: language derives meaning from its use~\cite{wittgenstein1958philosophical}. It is a communication tool humans develop to collaboratively solve embodied tasks, thus abstracting perception experiences like videos into symbolic concepts.
Emergent communication in machine learning~\cite{lazaridou2016multi, cangelosi2002computer,kirby2002natural,wagner2003progress,lazaridou2020emergent} aims to similarly learn an ``emergent language'' of perception in an unsupervised and domain-adaptive way: as Figure~\ref{fig:UniLearn frame}(a) shows, a speaker and a listener network play a referential game, where a speaker maps some visual stimuli (e.g.\,image, video) into a message of discrete tokens, and a listener uses the message to choose the speaker reference out of detractors. By jointly optimizing for game success, the communication protocol emerges structural and semantic properties resembling language, yet its discrete tokens might convey more fine-grained concepts than natural language useful for distinguishing stimuli. Thus, an emergent language can act as a bridge between natural language and videos, featuring a structure similar to the former while preserving the semantics of the latter.

Inspired by recent work that links emergent and natural language~\cite{yao2022linking}, we propose \textbf{E}mergent \textbf{C}ommunication for \textbf{E}mbodied \textbf{C}ontrol (\model), a three-phase scheme to learn embodied representations for downstream few-shot control (Figure \ref{fig:first graph}(a-c)).
First, we learn an emergent language of demonstration videos without using any language labels, which could potentially capture domain-specific concepts not directly available in paired natural language captions. 
Second, we learn an embodied representation by using a language model to predict trajectory segments conditional on trajectory contexts and language annotations, both natural and emergent. Rather than forcing video and language representations to be aligned, such a sequence modeling approach learns to represent natural language (for compositional generalization) and emergent language (for low-level perception and control) jointly with video trajectories, where the emergent language serves as a bridge between language and videos.
Third, we train a lightweight policy network on top of the embodied representation to few-shot adapt to downstream embodied control. 

We demonstrate that emergent language can help embodied control tasks and enable data-efficient few-shot downstream imitation learning via extensive experimental results across two existing benchmark simulation environments (Franka-Kitchen \cite{AbhishekGupta2019RelayPL} and MetaWorld \cite{yu2020meta}).   \model outperforms existing methods and achieves state-of-the-art performance on both language and video instruction following tasks. Our study also shows that although emergent language and natural language are utilized in a parallel manner in the pre-training phase, the word embeddings of natural language can still be predicted by linear regression models from emergent language. We further investigate the impact of emergent language and natural language correlation on the performance of downstream tasks and show that the performance in downstream tasks is best when the learned EC contains more information than natural language, i.e., when the accuracy of predicting natural language from emergent language is higher than that of predicting emergent language from natural language.
Overall, on the basis of these results, we believe that \model has great potential to become a  promising embodied control framework.

To sum up, our work makes the following contributions. (1) we build an embodied representation pre-training framework with emergent communication (EC) via a language model, which incorporates both the abstract, compositional structure of language and visuals with low-level motion details. (2) we develop a few-shot embodied control system, which transfers the pre-trained language model to downstream tasks as a frozen module and quickly adapts lightweight policy networks with a few expert data to generate actions. (3) we demonstrate that the learned representation with EC can benefit few-shot embodied control tasks and extensive experiments show that \model outperforms existing methods in Metaworld~\cite{yu2020meta}  and Franka Kitchen~\cite{gupta2019relay} benchmarks.

\section{Related Work}
\subsection{Emergent Communication}
The recent advances in NLP ~\cite{devlin2018bert,radford2019language,brown2020language} benefits from huge text corpora and are striving to find statistical regularities on large amounts of data via big models like GPT \cite{radford2018improving,radford2019language,brown2020language} and BERT \cite{devlin2018bert}. This narrow focus was leaving aside language emergence and language grounding (in some other modalities, e.g., sight, with more complex inputs than text). In contrast, human language acquisition ~\cite{de1978language} starts with a cumulative series of interactions with other people ~\cite{bruner1985role,barton1994input} grounded in the physical and social world ~\cite{harnad1990symbol,vogt2002physical,alomari2017natural} and does not rely on passive and complex corpora like Wikipedia. Emergent Communication (EC) is one promising direction toward this motivation, where a communication protocol is shaped through multi-agent interactions with perceptual grounding like a human\cite{patel2021interpretation}. As a typical example, the referential image game involves a speaker creating a sequence of discrete tokens based on an input image, and a listener tasked with selecting the correct input from a group of distractors using the message. Both networks are optimized jointly using success signals from the game\cite{yao2022linking}. By studying games like this type, researchers are interested in the emergence of desirable properties resembling natural language, such as game success generalization and compositionality, which holds great promises towards more functional ~\cite{wittgenstein1958philosophical,wagner2003progress} and generalizable ~\cite{lake2017building} language agents.

\subsection{Embodied Control under Instruction}
Embodied control has achieved much success in learning grasping and pick-place tasks from low-dimensional state~\cite{argall2009survey,khansariTRO2011,billard2004discovering,schaal2005learning,chalodhorn2007learning,pastor2009learning,mulling2013learning}. Deep learning makes embodied control directly learn from high dimensional observations becoming feasible~\cite{pomerleau1989alvinn,zhang2018deep, rahmatizadeh2018vision}.
Recent advances aim to build a learning-based embodied control system with deep neural networks that are flexibly conditioned on either a demonstration video ~\cite{yu2018one,bonardi2020learning} of a human or a language instruction ~\cite{stepputtis2020language, lynch2020grounding} via multi-modal representation learning. With the guidance of instruction, the embodied agent aims to generalize to new  scenes~\cite{pathak2018zero}, novel objects~\cite{finn2017one,james2018task,yu2018one,bonardi2020learning,zhou2019watch}, novel object configurations~\cite{paine2018one}, and novel goal configurations~\cite{duan2017one,huang2019neural,dasari2020transformers}. For instance, BC-Z~\cite{jang2022bc} has developed a flexible imitation learning system that can learn from demonstrations and interventions, based on various forms of task information, such as pre-trained embeddings of natural language or human performance videos. R3M~\cite{nair2022r3m} pre-trains visual representations on diverse human video-language data by encouraging alignment between the two modalities. However, since videos often convey more detailed information than words, this framework may lose valuable teaching information that can help robots perform complex tasks.
In this work, we aim to build an instruction-conditioned embodied control framework without forcing alignment on video and language and consider the video demonstrations and language instructions as “parallel sentences” during the pre-training process to learn useful information for downstream tasks.

\section{Method}
As shown in Figure \ref{fig:UniLearn frame},  we first pre-train a speaker and listener via an emergent communication game for emergent language generation and a language model to extract embodied representations by predicting the cropped part of a latent trajectory. The language model takes the emergent language or natural language as prompt and the masked trajectory as input and aims to predict the masked part of the trajectory separately. Then, we transfer the learned speaker and language model into downstream embodied control tasks as frozen modules. The output of the language model extracts embodied representation from instruction prompt and current observations and is mapped into actions by the downstream policy network, which contains a few task-specific  MLP~\cite{ramchoun2016multilayer} layers.
\subsection{Problem Statement}
The goal for embodied control is to learn a policy $\pi_{B}$ conditioned on the instruction, either language instruction or video instruction, that reproduces the expert behavior on the desired task specified by the instruction. Usually, the demonstrations of experts are presented in the form of state-action trajectories, with each pair indicating the action to take at the state being visited. 
 Let $\tau_E$ denote a trajectory sampled from expert policy $\pi_E(s|c)$: $\tau_E = \big[ (s_0, a_0), (s_1, a_1), \ldots, (s_n, a_n) \big]$. 
 The instruction is encoded into the latent vector $c$. To learn the behavior policy $\pi_{B}(s|c)$, the demonstrated actions are usually utilized as the target label for each state, and the mapping $a \in \pi_{B}(s|c)$ from states to actions is learned in a supervised manner with given instructions.

\subsection{Emergent Communication Language Generation for demonstration video }
\label{sec:ec}
Our idea is to automatically generate synthetic language that aligns with video demonstration through emergent communications, which connect perceptual, language, and action for pre-training. 
As shown in Figure~\ref{fig:UniLearn frame}(c), we consider a typical speaker-listener referential game~\cite{lazaridou2016multi,lazaridou2018emergence,li2019ease} on a set of $N$ video features $\mathcal{D}_I = \{ I_1, \cdots, I_N \}$. At each training step, the speaker takes one video feature $I_i$ and generates a discrete message $M_i \in [V]^{T}$ using the Gumbel-Softmax trick~\cite{jang2016categorical}, where $V$ is the vocabulary size, and $T$ is the sequence length limit. For simplicity, denote $m=M_i$ so that $m_t = M_{i, t}$ denotes the $t$-th token of the message $M_i$. The generated discrete message is defined as ``emergent language", and the process of emergent language generation by the speaker can be formulated as
\begin{equation}
\begin{aligned}
     \textbf{hs}_0 = I_i,  \quad \textbf{hs}_t &= \mathrm{GPT}_{\text{spk}}\left(m_{t-1}, \textbf{hs}_{t-1}\right) \ (t>0), \\
     m_0 = \text{[CLS]}, \quad  m_t &= \text{Softmax}\left(\mathrm{MLP}_{\text{spk}}( \textbf{hs}_t)\right) \ (t>0).
\end{aligned}
\label{eq:spk}
\end{equation}
Here $\textbf{hs}_t$ denotes speaker hidden states. Next, the listener takes the message $m$, and tries to guess the right video $I_i$ out of a set of $K$ confounding videos $C_i = \{I_{j_1}, \cdots, I_{j_K}\} \subset \mathcal{D}_I - \{I_i\}$. Note that different from the EC games on images, in order to learn temporal information, except for irrelevant videos, we also construct several additional candidates in $C_{i}$ by performing temporal augmentation on the original video, such as reverse order and random disorder.
The listener also uses a GRU~\cite{chung2014empirical} layer to turn the message $m$ into a hidden vector $\textbf{hl}_{T}$ 
\begin{align}
    \mathbf{hl}_0 = \mathbf{0}, \quad \mathbf{hl}_t &= \mathrm{GPT}_{\text{lsn}}\left(m_{t}, \mathbf{hl}_{t-1}\right) \ (t>0).
\end{align}
Based on $\mathbf{hl}_{T}$, the listener assigns a score for each candidate video based on inverse square error~\cite{lee2017emergent}, then selects the image by Softmax sampling across the scores. The speaker and listener are jointly optimized by minimizing the cross-entropy loss of video selection:
\begin{equation}
\begin{aligned}
\operatorname{score}(\mathbf{I}) &=\left\|\mathbf{h l}_T-\text{MLP}_{\text{lsn}}(\mathbf{I})\right\|_2^{-2}, \\
p(\text { guess }=\mathbf{I}) &=\operatorname{softmax}(\operatorname{score}(\mathbf{I})) \quad\left(\mathbf{I} \in\left\{\mathbf{I}_i\right\} \cup C_i\right), \\
\mathcal{L}_{E C} &=-\mathbb{E}_{\mathbf{I}_i, C_i} \mathbb{E}_{\mathbf{M}_i} \log p\left(\text { guess }=\mathbf{I}_i\right) .
\end{aligned}
\end{equation}
The speaker can be employed to generate emergent language $\mathcal{D}_{M} = \{ M_1, \cdots, M_N\}$ based on input videos. 
\subsection{Pre-training of Language Model with Emergent Language}

We conduct a trajectory completion task to pre-train the GPT-like language model without action labels, as shown in Figure \ref{fig:UniLearn frame}(a)(b). Firstly, we random sample a sequence of observations $\hat{o}=\{o_{1},o_{2}, \ldots,o_{N}\}$ stored in the dataset and map it into latent trajectory $\tau_{\text{whole}}$ by encoder $g_{\theta}$.
\begin{equation}
 \tau_{\text{whole}}=g_{\theta}(\hat{o})   
\end{equation}
Then we crop a random segment $\tau_{\text{seg}}$ from the whole latent trajectory $\tau_{\text{whole}}$ and the remained trajectory is denoted as $\tau_{\text{rem}}$.
The language model $f_{\phi}$ takes $\tau_{\text{rem}}$ as input and uses either emergent language $e$ generated by the speaker or natural language $l$ as a prompt to predict the cropped segment $\tau_{\text{seg}}$. The language model is optimized by jointly minimizing $L_{\text{EC}}$ and $L_{\text{Lang}}$
\begin{equation}
 L_{\text{EC}}=(\tau_{\text{seg}}-f_{\theta}(e,\tau_{\text{rem}}))^{2} 
\end{equation}
\begin{equation}
L_{\text{Lang}}=(\tau_{\text{seg}}-f_{\theta}(l,\tau_{\text{rem}}))^{2}   
\end{equation}
The loss for the prediction using emergent language as prompt and natural language as prompt is calculated separately. 

\model uses the same language model for both emergent language generation and trajectory completion pretext task for framework thriftiness, and both the $L_{\text{EC-pre}}$ and $L_{\text{EC-gen}}$ are used to update the language model jointly. To ensure that the speaker's gradient is not truncated, we use an MLP~\cite{ramchoun2016multilayer} layer to map from the logits of the word distribution to a 512-dimensional latent vector instead of selecting a word embedding in the dictionary according to the token. For natural language, we use the Byte-Pair Encoding (BPE) tokenizer \cite{bostrom2020byte} to encode it into tokens, which are widely used by OpenAI for tokenization when pre-training the GPT model, including GPT \cite{radford2018improving}, GPT-2\cite{radford2019language}, RoBERTa\cite{liu2019roberta}, and BART\cite{lewis2019bart}. We also learn a corresponding dictionary of word embedding. Although the emergent language and natural language are used separately, they can learn correlated information via the pretext trajectory completion task.

\subsection{Few-shot Policy Learning  for Downstream Embodied control}
To transfer the pre-trained language model into the downstream imitation learning tasks, the language model $f_{\theta}(\cdot,\cdot)$ take the current trajectory $\tau_{cur}=\{s_{t-T},\ldots,s_{t-1},s_{t}\}$ and the emergent language $e$ or natural language instruction $l$ as prompt, then output the predicted features $m_{t}$. Finally, the output $m_{t}$ of the language model is mapped to specific action $a_{t}$ by task-specific MLP layers $\mathcal{\pi}_{\psi}$.  
The language model $f_{\theta}$ is frozen during downstream imitation learning.
The action under learned behavior policy is sampled by $\hat{a}_{t}\sim\pi_{\psi}(a_{t}|m_{t})$ and the expert demonstration action $a_{t}$ are used as the label for imitation learning. For the video instruction following task, we generate emergent language by the speaker and use it as the prompt of the language model. For language instruction following tasks, the natural language is tokenized by the BPE tokenizer, and select the corresponding word embedding in the dictionary as the prompt as language model via the inferred token. With a trajectory with length $H$, the task-specific MLP layers are optimized by minimizing $\mathcal{L}_{\text{IL}}(\psi)$.

\begin{equation}
\footnotesize
{
 \mathcal{L}_{\text{IL}}(\psi)=E_{\tau \sim \mathcal{B}}\left[-\frac{1}{H} \sum_{t}^{t+H-1} \log \pi_{\psi}(a_{t}|m_{t})\right]  }
\end{equation}
The MLP~\cite{ramchoun2016multilayer} layers are implemented with hidden sizes [256, 256] with batch normalization~\cite{ioffe2015batch} and are trained with a learning rate of 0.001 and a batch size of 32 for 20000 iterations.
\section{Experiments}
\subsection{Pre-training Dataset}
We use the LOReL\cite{nair2022learning} real robot dataset, which is collected by Franka Emika Panda mounted over an IKEA desk. The LOReL dataset contains 3000 episodes (150000 frames) of reinforcement learning that trains policies for different behaviors on the IKEA desk using online RL.
The language instruction is annotated by leveraging crowd-sourcing, specifically Amazon Mechanical Turk. 
Human annotators are asked to describe the behavior if any, that the robot is doing and to phrase it as a command without any pre-specified template. 
The LOReL dataset contains 6000 annotations, two per episode, containing a total of 1699 unique instructions. The dataset contains different videos with the same language instruction to ensure diversity. 
We removed the episodes where annotators reported the robot as inactive or its actions unclear, making them incomprehensible.
 \begin{figure*}[t]
    \centering
    \includegraphics[width=0.88\linewidth]{ 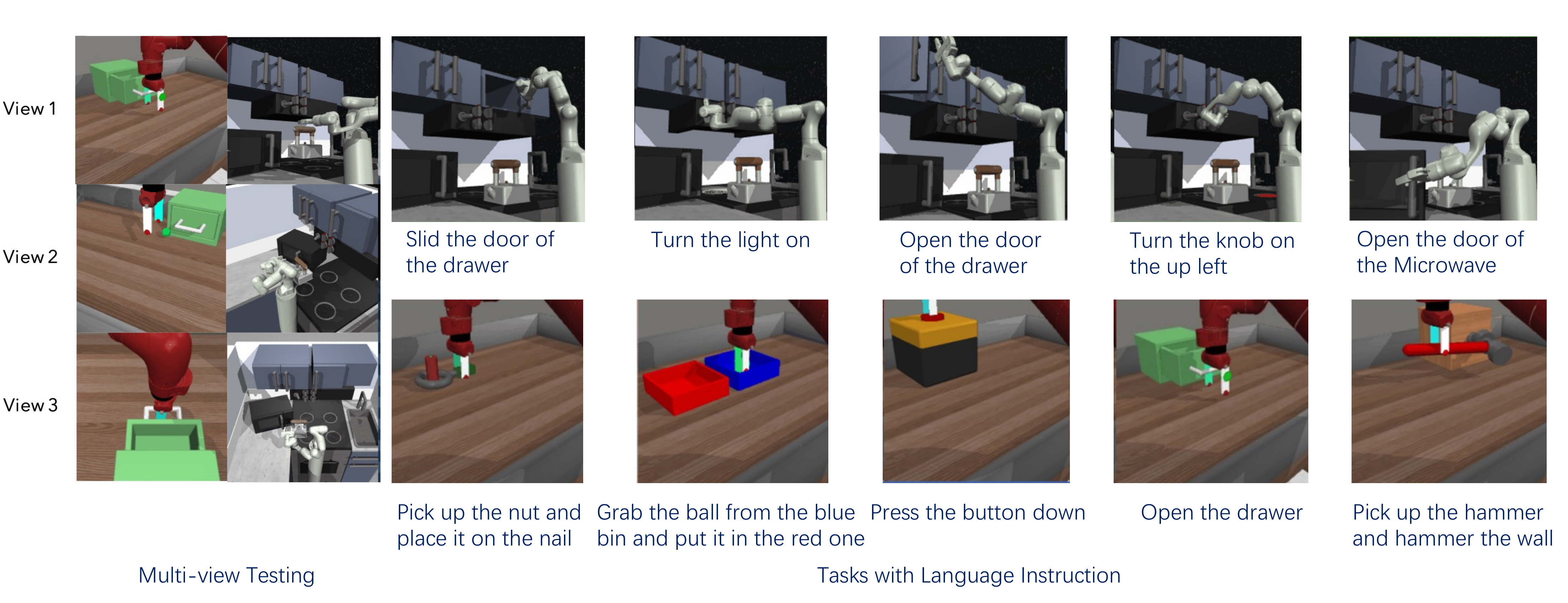}
    \vspace{-10pt}
    \caption{ \textbf{Evaluation Environments}: We consider a set of manipulation tasks including 5  tasks with a Sawyer from MetaWorld \cite{yu2020meta}, 5 tasks from a Franka operating over a Kitchen \cite{gupta2019relay}. We provide specific language instruction labels for every task and 25 demos per task for evaluation. We show an example image and corresponding language instruction for each task shown above.
    \textbf{Evaluation Viewpoints}: To robustly measure the efficacy of different visual representations, we consider three camera viewpoints per environment.
    }
    \vspace{-5pt}
    \label{fig:Evaluation Environments}
\end{figure*}
\begin{table*}[t]
\scriptsize
\centering
\resizebox{\linewidth}{!}{
\begin{tabular}{c|ccllll}
\toprule
        Benchmark & Num. Demos& Instruction         &\model (ours) &  R3M  ~\cite{nair2022r3m} & BC-Z ~\cite{jang2022bc} \\
\hline
\multirow{4}{*}{MetaWorld}&\multirow{2}{*}{10}& Lang& $\textbf{72.4}\pm{2.4\%}$ &$69.2\pm{2.0\%}$ & $63.4\pm{2.1\%}$   \\
&&\cellcolor{codegray}Video & \cellcolor{codegray}$\textbf{76.0}\pm{2.1\%}$ (\textcolor{blue}{+\textbf{3.6}})& \cellcolor{codegray}$71.8\pm{1.8\%}$ (\textcolor{blue}  {+2.6}) & \cellcolor{codegray}$64.8\pm{2.0\%}$ (\textcolor{blue}  {+1.4})  \\
\cline{2-6}
&\multirow{2}{*}{25}& Lang& $\textbf{82.6}\pm{2.0\%}$ & $77.4\pm{2.4\%}$ & $74.1\pm{2.2\%}$ \\
&&\cellcolor{codegray}Video &\cellcolor{codegray}$\textbf{85.0}\pm{2.2\%}$ (\textcolor{blue}{+\textbf{2.4}})& \cellcolor{codegray}$79.0\pm{2.1\%}$ (\textcolor{blue}  {+1.6}) & \cellcolor{codegray}$75.2\pm{1.0\%} (\textcolor{blue}{+1.1})$\\
\midrule
\multirow{4}{*}{Franka Kitchen}&\multirow{2}{*}{10} &Lang  & $\textbf{48.4}\pm{2.5\%}$ & $41.8\pm{2.5\%}$ & $34.2\pm{2.3\%}$  \\
&&\cellcolor{codegray}Video & \cellcolor{codegray}$\textbf{53.6}\pm{2.5\%}$(\textcolor{blue}{+\textbf{5.2}}) & \cellcolor{codegray}$45.7\pm{2.4\%}$ (\textcolor{blue}  {+3.9}) & \cellcolor{codegray}$37.5\pm{2.2\%}$ (\textcolor{blue}{+3.3})\\
\cline{2-6}
&\multirow{2}{*}{25}&Lang  & $\textbf{59.8} \pm{2.5\%}$ & $56.0\pm{2.3\%}$ & $38.6 \pm{2.4\%}$\\
&&\cellcolor{codegray}Video& \cellcolor{codegray}$\textbf{63.2}\pm{2.6\%}$ (\textcolor{blue}{+\textbf{3.4}})& \cellcolor{codegray}$58.7\pm{2.0\%}$ (\textcolor{blue}  {+2.7})& \cellcolor{codegray}$40.0\pm{2.0\%}$ (\textcolor{blue}{+1.4})\\
\toprule
\end{tabular}}
\vspace{-10pt}
\caption{Performance comparison on Success rate of embodied control with language/video instruction across different demo size. We mark the performance difference between complete the task with  natural language instruction and video instruction as blue.}
\vspace{-15pt}
\label{tab:main_result}
\end{table*}
\begin{table*}[t]
 \footnotesize
\centering
\resizebox{\linewidth}{!}{
\begin{tabular}{l|c|cc|ccc}
\toprule
\multirow{2}{*}{Benchmark}&\multirow{2}{*}{Num. Demos}& \multicolumn{2}{c|}{Language Instruction} & \multicolumn{3}{c}{Video Instruction} \\
\cline{3-7}
             &&\model & \model(-EC) & \model  & \model(-NL) & \model(+Cap,-EC) \\
\hline
\multirow{2}{*}{MetaWorld}&10& $\textbf{72.4}\pm{2.4\%}$ & $69.0\pm{2.0\%}$ & $\textbf{76.0} \pm{2.1\%}$ & $72.6\pm{2.4\%}$ & $73.0\pm{2.2\%}$ \\
&25 & $\textbf{82.6}\pm{2.0\%}$ & $78.0\pm{2.4\%}$ & $\textbf{85.0} \pm{2.5\%}$ & $82.0\pm{2.0\%}$ & $73.2\pm{2.5\%}$ \\
\hline
\multirow{2}{*}{Franka Kitchen}&10 & $\textbf{48.4}\pm{2.5\%}$ & $43.0\pm{3.3\%}$ & $\textbf{53.6}\pm{2.5\%}$ & $49.0\pm{2.5\%}$ & $48.6\pm{2.4\%}$\\
&25 & $\textbf{59.8}\pm{2.5\%}$ & $54.0\pm{2.6\%}$  & $\textbf{63.2}\pm{2.6\%}$ & $61.3\pm{2.4\%}$ & $49.0\pm{2.3\%}$ \\
\toprule
\end{tabular}}
\vspace{-8pt}
\caption{Ablation study on embodied control with language instruction and video instruction.}
\vspace{-5pt}
\label{tab:ablation-study_whole}
\end{table*}
\subsection{Downstream Testing Environments}
We utilize two standard robotic manipulation benchmarks, namely MetaWorld \cite{yu2020meta} and the Franka Kitchen environment \cite{gupta2019relay}, as our downstream testing environments. It is worth noting that these testing environments were not used during the training phase of \model. In MetaWorld, the agent is presented with a series of tasks, including assembling a ring on a peg, picking and placing a block between bins, pushing a button, opening a drawer, and hammering a nail. In Franka Kitchen, the agent aims to learn how to slide open the right door, open the left door, turn on the light, turn the stove top knob, and open the microwave. As shown in Figure \ref{fig:Evaluation Environments}, both environments offer image observations to the agent, as well as proprioceptive data, which includes the end-effector pose and joint positions. These data are concatenated with the encoded vision observations for each task, providing the agent with a comprehensive understanding of the current task.
 All tasks involve random environment variation, either by varying the position of the target object in MetaWorld or the positioning of the desk in Franka Kitchen. Additionally, we consider three views for each environment to evaluate the robustness of representations across viewpoints. Finally, for each environment, we also consider three demo sizes used in downstream embodied control tasks: $[5, 10,25]$ in MetaWorld and Franka Kitchen. 

\subsection{Experimental Setup}
For embodied control tasks, we take R3M and BC-Z as the key baseline. \textbf{R3M} ~\cite{nair2022r3m} learns the state-of-the-art embodied representations pre-trained on diverse paired human video-language data with time-contrastive learning, which encourages states closer
in time to be closer in embedding space and video-language alignment to encourage the embeddings to capture
semantically relevant features. For sufficient comparison, we implement R3M with both the language instruction-conditioned version and video instruction-conditioned version, respectively, based on their official implementation. For embodied control with natural language instruction, we use the language encoder of the pre-trained model officially released by their author team to encode the natural language. For embodied control with demonstration video instruction, we use the officially released vision encoder to extract features of every frame in the video and map them into the instruction representation by one MLP layer. \textbf{BC-Z} ~\cite{jang2022bc} jointly learns the alignment between video representation and language representation together with policy learning across multiple embodied control tasks.
Our evaluation methodology is inspired by ~\cite{nair2022r3m, SimoneParisi2022TheE}. We focus on evaluating the embodied representation of instruction and observation for downstream policy learning with behavior cloning. We parameterize the downstream policy  $\pi$ as a two-layer MLP preceded by a BatchNorm at the input. We train the agent for 20,000 steps, evaluate it online in the environment every 1000 steps, and report the best success rate achieved. For each task, we run 5 seeds of behavior cloning. The final success rate reported for one method on one task is the average over the 5 seeds $\times$ 3 camera viewpoints $\times$ 3 demo sizes, for a total of 45 runs (see more details in Appendix).

\begin{table*}[t]
\begin{minipage}[b]{0.3\textwidth}
\resizebox{\linewidth}{!}{
\renewcommand\arraystretch{1.1}
\setlength{\tabcolsep}{3pt}
\begin{tabular}{cccc}
\toprule
Model Size & \#layer &  \#head & \#embded \\
\hline
Base     & 8          & 16        & 512         \\
Mini     & 6          & 6         & 192         \\
Micro    & 4          & 4         & 128 \\
\toprule
\end{tabular}}
\vspace{-4pt}
\caption{Hyper-parameters for different model sizes of \model.
All the models are implemented based on GPT~\cite{radford2018improving}-like network architecture.
}
\label{tab:model size}
\end{minipage}
\hfill
\begin{minipage}[b]{0.68\textwidth}
\footnotesize
\resizebox{\linewidth}{!}{
\begin{tabular}{l|l}
\toprule
Natural Language      & Emergent language \\
\hline
``Open the right \textcolor{drawercolor}{\textbf{drawer}}” & {[}262, 847, 490, 536, \textcolor{drawercolor}{\textbf{108}}, \textcolor{drawercolor}{\textbf{500}}, 941, 838, 749, 444{]}\\
``Grasp in the middle \textcolor{drawercolor}{\textbf{drawer}}"&{[}356,  69, \textcolor{drawercolor}{\textbf{108}}, 259,  72, 616, \textcolor{drawercolor}{\textbf{500}}, 896, 623, 433{]}\\
``\textcolor{reachcolor}{\textbf{Reach}} down for \textcolor{staplercolor}{\textbf{stapler}}" & {[}204, 897, 285, 509, \textcolor{staplercolor}{\textbf{701}}, 545, \textcolor{reachcolor}{\textbf{538}}, 729, 902, 195{]}\\
``Pick up the \textcolor{staplercolor}{\textbf{stapler}} in the \textcolor{drawercolor}{\textbf{drawer}}" &{[}535, 67, \textcolor{drawercolor}{\textbf{108}}, 661, 84, \textcolor{staplercolor}{\textbf{701}}, 776, \textcolor{drawercolor}{\textbf{500}}, 377, 454{]}\\
``\textcolor{reachcolor}{\textbf{Reach}} to the plug" &{[}982,  691,   72,  \textcolor{reachcolor}{\textbf{538}},   97,  665,  148,   369,   102, 1003{]}\\
``\textcolor{reachcolor}{\textbf{Reach}}  to the left \textcolor{drawercolor}{\textbf{drawer}}" & {[}786, \textcolor{reachcolor}{\textbf{538}}, 601, 874, 848, \textcolor{drawercolor}{\textbf{500}}, 315, \textcolor{drawercolor}{\textbf{108}}, 72, 914{]}\\
\toprule
\end{tabular}}
\vspace{-10pt}
\caption{Qualitative examples of emergent language and corresponding natural language. The word and tokens that have the same semantic meaning are marked with the same color.}
\label{tab:Qualitative examples}
\end{minipage}
\vspace{-15pt}
\end{table*}

\subsection{Results}

\begin{figure*}[t]
  \centering
    \begin{subfigure}{0.245\textwidth}
      \centering   
      \includegraphics[width=\linewidth]{ 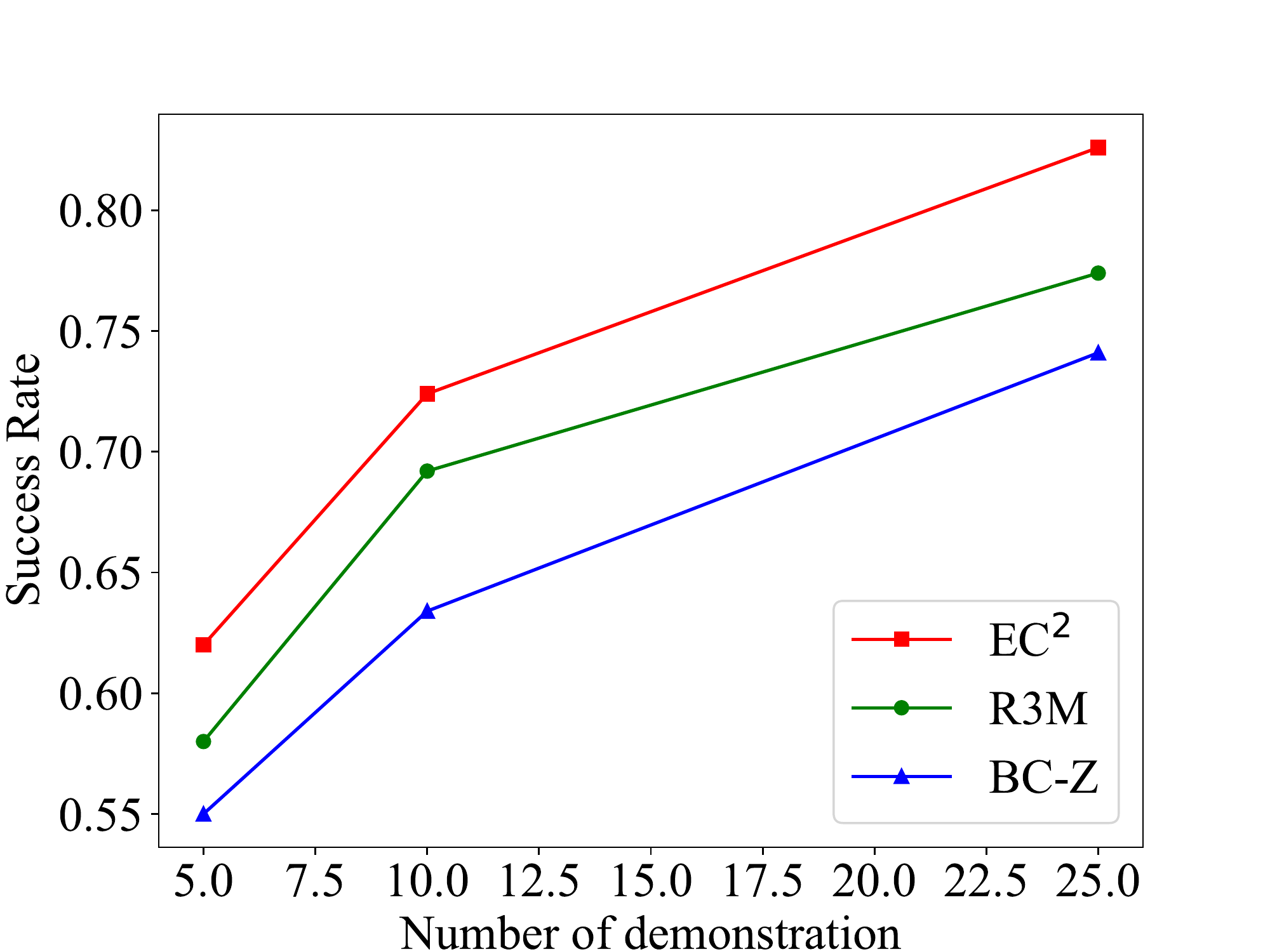}
        \caption{MetaWorld[\textit{\textbf{language}} instruction]}
        \label{fig:sub1}
    \end{subfigure}  
    \begin{subfigure}{0.245\textwidth}
      \centering   
      \includegraphics[width=\linewidth]{ 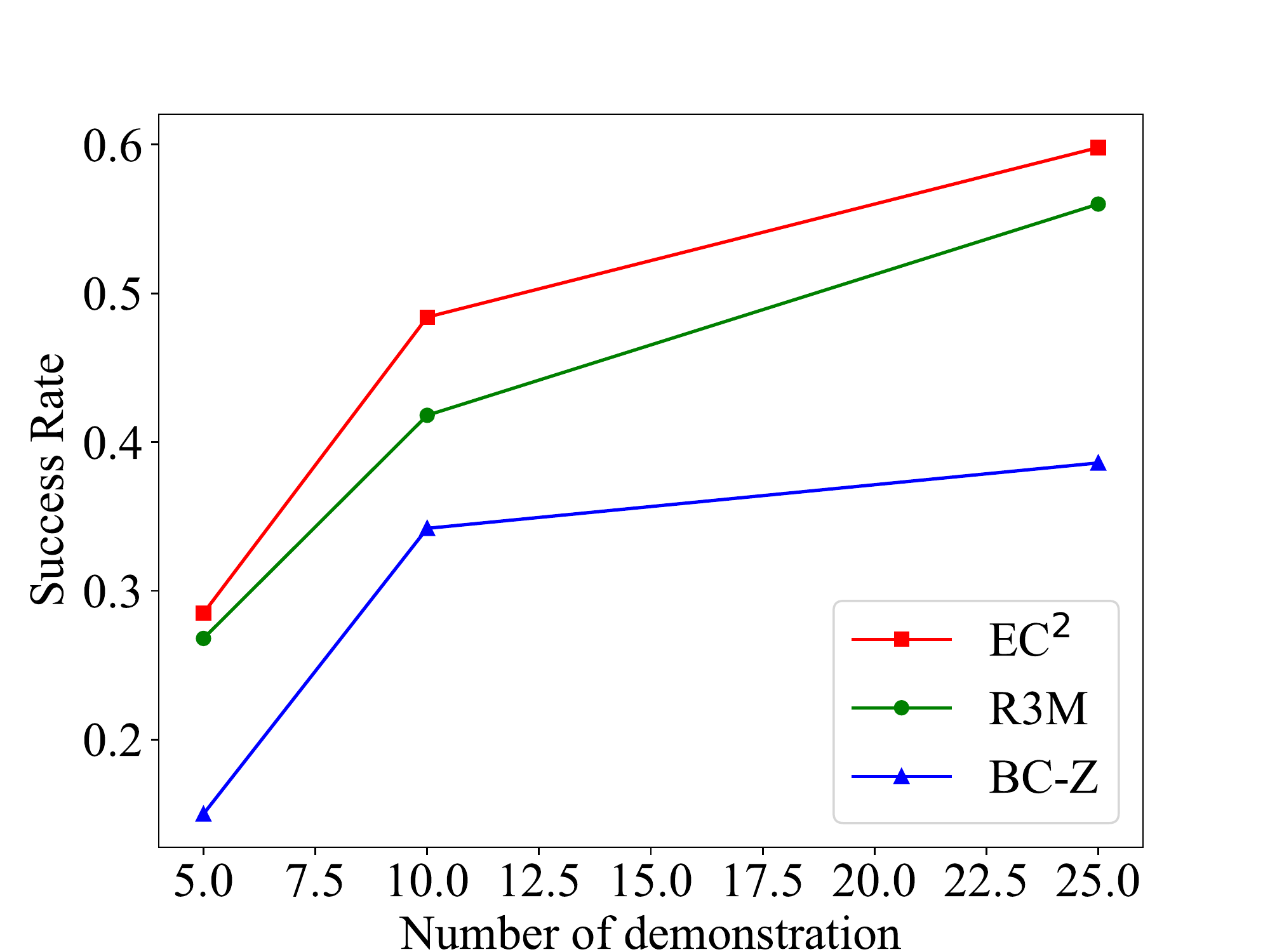}
        \caption{Kitchen[\textit{\textbf{language}} instruction]}
        \label{fig:sub2}
    \end{subfigure}
    \begin{subfigure}{0.245\textwidth}
      \centering   
      \includegraphics[width=\linewidth]{ 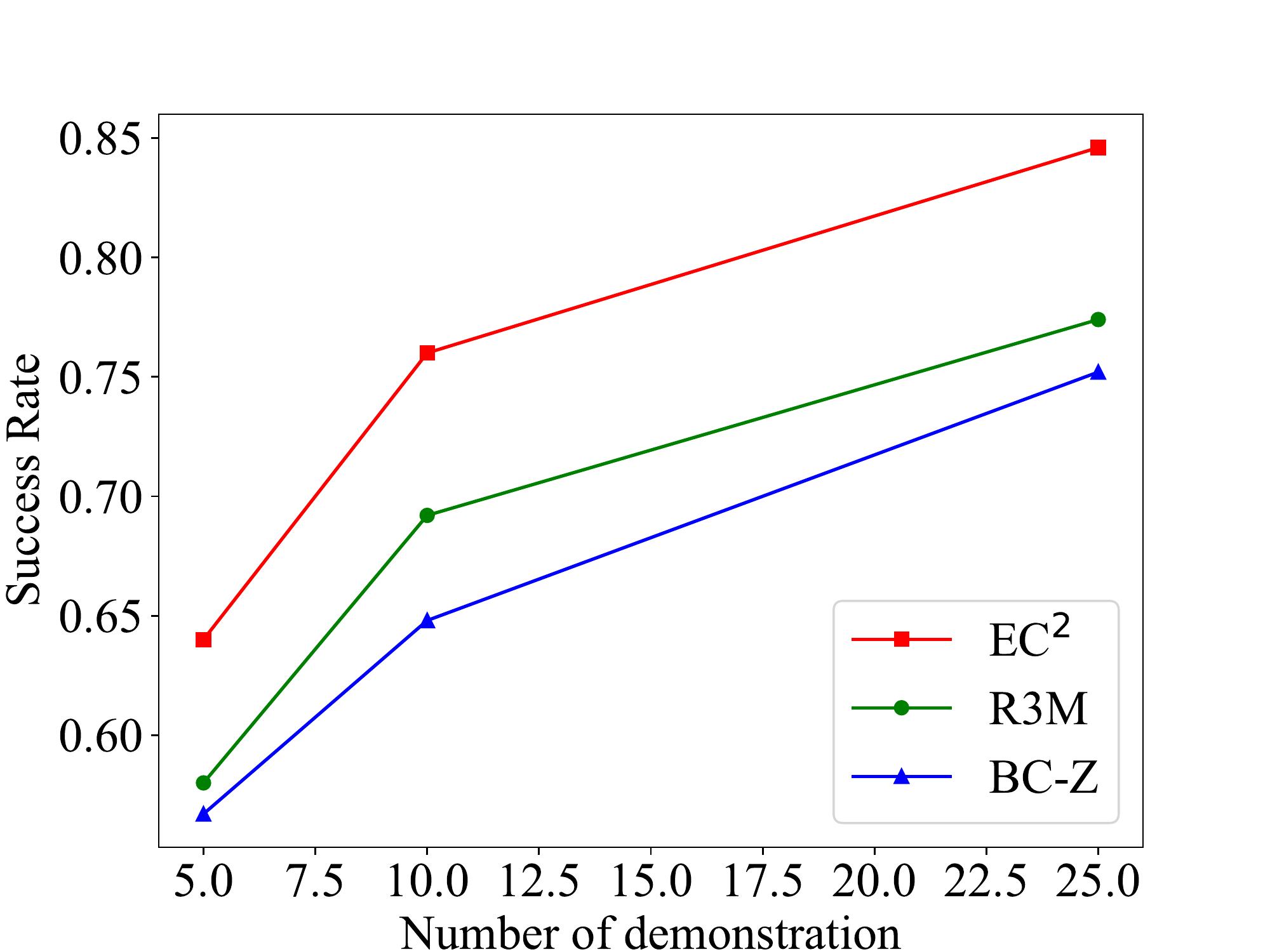}
        \caption{MetaWorld[\textit{\textbf{video}} instruction]}
        \label{fig:sub3}
    \end{subfigure}
    \begin{subfigure}{0.245\textwidth}
      \centering   
      \includegraphics[width=\linewidth]{ 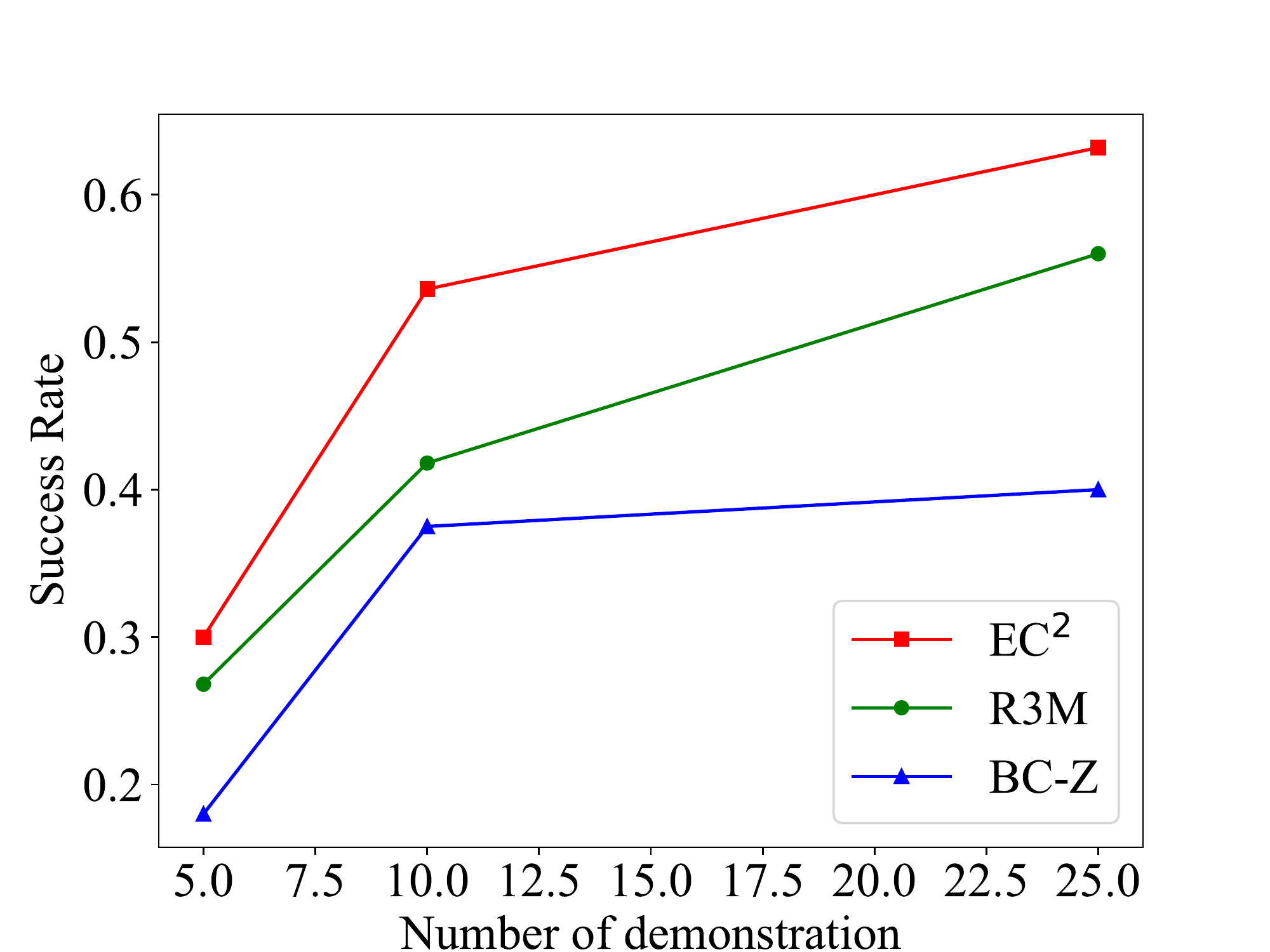}
        \caption{Kitchen[\textit{\textbf{video}} instruction]}
        \label{fig:sub4}
    \end{subfigure}
    \vspace{-8pt}
    \caption{Performance over different demo sizes used in downstream imitation learning. We report the success rate of \model and baseline across different demo sizes. We see that the performance improvement from \model is consistent across all demo sizes.
    }
    \vspace{-15pt}
    \label{fig: different dataset}
\end{figure*}

We aim to answer 5 questions to get insight into embodied AI community:
1) Can emergent language help embodied control tasks with natural language instruction?
2) Can emergent language help the embodied control tasks with video instruction?
3) Can the joint training of emergent language and natural language in GPT-Like network architecture help each other? 
4) Can emergent language provide more detailed and effective guidelines than video captions?
5) What is the relationship between model size, control performance, and the correlation between emergent language and natural language?

To answer the first two questions, we study if \model enables more data-efficient
policy learning with both natural language instructions and video instructions.

1) \textit{Can emergent language help the embodied control tasks with natural language?}

We measure the success rate of downstream few-shot policy learning with different pre-training methods. In Franka Kitchen and MetaWorld, we only use no more than 25 demos to perform downstream few-shot policy learning and report the performance with both 10 and 25 demos separately in Table \ref{tab:main_result}. Learning from scratch is struggled to perform well when working with small amounts of data since it can be challenging for the model to identify meaningful  patterns or relationships with such limited data.  While the methods with a pre-trained model, like \model, R3M, and BC-Z performs better than those learning from scratch. Through our study, as shown in Table \ref{tab:main_result}, \model exceeds existing methods with natural language instruction, which indicates that emergent language can help the embodied control tasks to improve the performance with natural language instruction. Across all the evaluations, \model is overall able to learn these embodied tasks in an extremely low data regime (only 10 demos) with $\approx$60.4\% overall success rate, despite never seeing any data from the target environments in training the representation, while R3M only achieves $\approx$55.5\% overall success rate.

\begin{figure*}[t]
     \centering
          \begin{subfigure}[b]{0.33\textwidth}
         \centering
         \includegraphics[width=\textwidth]{ 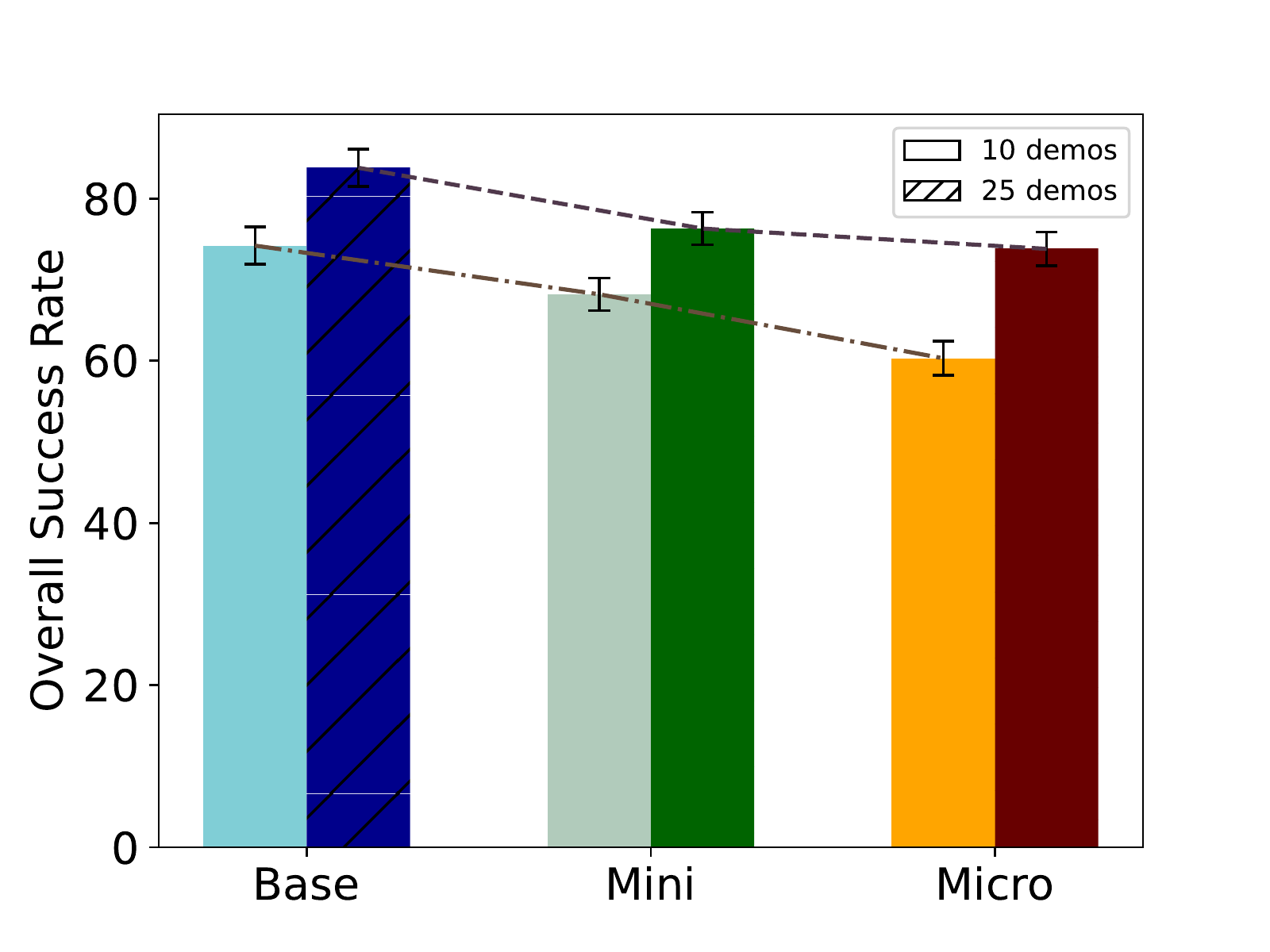}
         \vspace{-15pt}
         \caption{Ablation on model size in MetaWorld}
         \label{fig:ablation_model_meta}
     \end{subfigure}
     \begin{subfigure}[b]{0.33\textwidth}
         \centering
         \includegraphics[width=\textwidth]{ 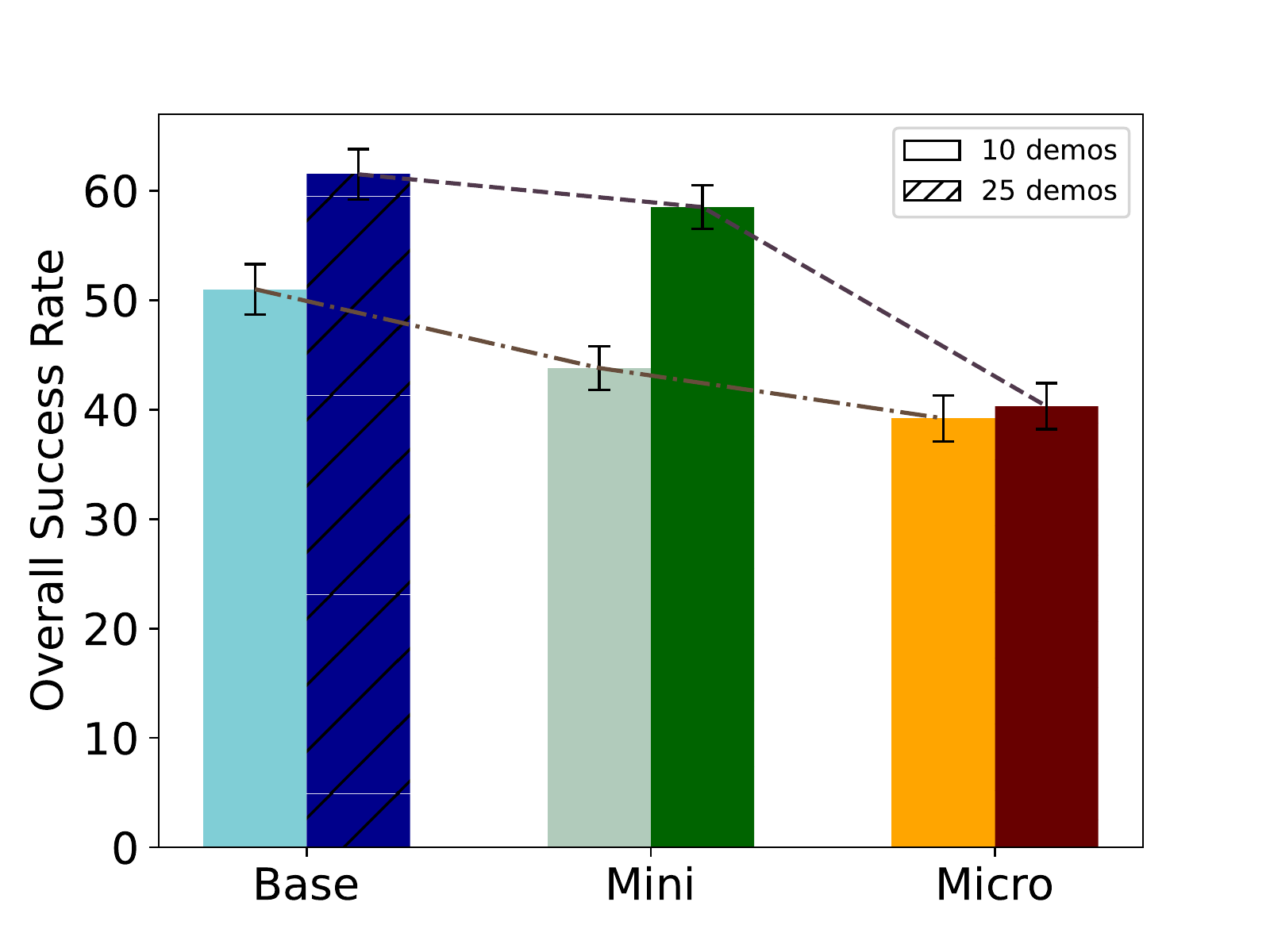}
         \vspace{-15pt}
         \caption{Ablation on model size in Franka Kitchen}
         \label{fig:ablation_model_kitchen}
     \end{subfigure}
     \begin{subfigure}[b]{0.33\textwidth}
         \centering
         \includegraphics[width=\textwidth]{ 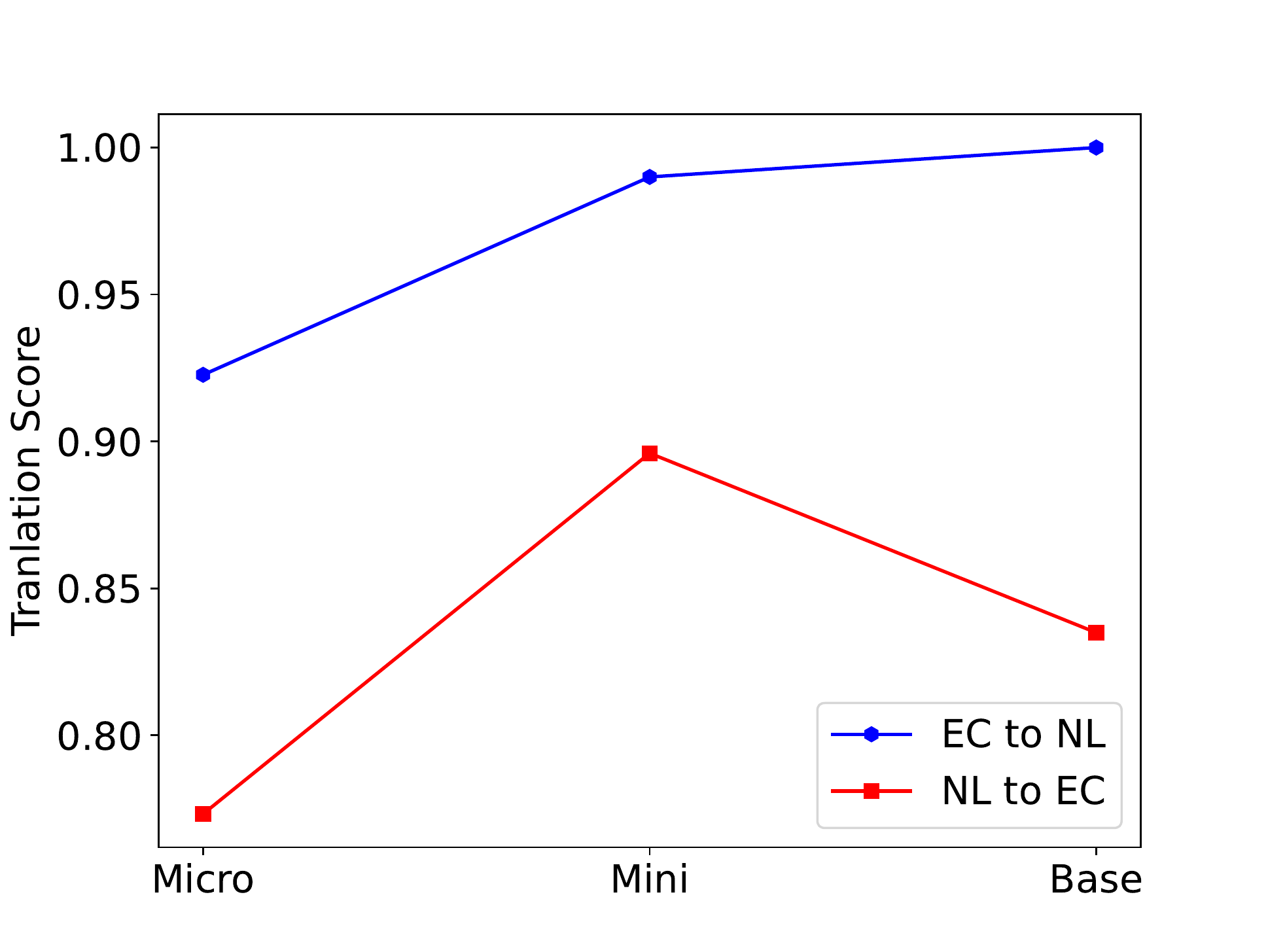}
         \vspace{-15pt}
         \caption{Translation score across different model sizes}
         \label{fig:understand}
     \end{subfigure}
     \vspace{-18pt}
        \caption{Ablation on model size and its relationship with the correlation of emergent language and Natural language. We show the ablation results on model size with 10 demos and 25 demos in downstream policy learning separately. }
        \vspace{-15pt}
        \label{fig:corelation}
\end{figure*}
2) \textit{Can emergent language help the embodied control tasks with video instruction?}

As shown in Table\ref{tab:main_result}, \model shows an obvious advantage over the exciting methods and achieves the best success rate ($\approx$64.8\% with 10 demos and $\approx$74.1\% with 25demos) with video instruction. As marked in blue in Table \ref{tab:main_result}, \model has better performance with video instruction than with language instruction, which indicates that \model learns more effective information provided by a video demonstration. In contrast, R3M and BC-Z, which use contrast learning or regression to force the two modalities of language and video aligned, have little difference in performance under language instruction and video instruction.

We also investigate the task performance of \model and baselines with different amounts of demos in policy learning. In Figure \ref{fig: different dataset}, we plot the average success rate of each method across each demo size. We observe that the performance improvement from \model is consistent, outperforming the baselines across every environment and demo size.

\noindent \textbf{Ablation Study}:
Also, in the data-efficient imitation learning setting, we ablate the different components of \model training to answer question 3 and 4 and check if each module is indispensable.

3) \textit{Can the joint training of emergent language and natural language with GPT-Like network  help each other? }

We conduct ablation studies to compare \model and \model~(-EC) in language instruction following tasks and compare \model and \model~(-NL) in video instruction following tasks. \model~(-EC) does not use emergent language to pre-train the language model under \model framework, which predicts the masked trajectories with only the language instruction (human-labeled caption).  \model~(-NL) does not use natural language to pre-train the language model under \model framework, which predicts the masked trajectories with only the emergent language.
As shown in the left-hand column in Table \ref{tab:ablation-study_whole}, without the help of emergent language, the performance decreased significantly, and \model~(-EC) shows no obvious advantages to the other methods that learn pre-trained models by video-language pre-training. Similarly,  as shown in the right-hand column in Table \ref{tab:ablation-study_whole}, \model also outperforms \model~(-NL), which demonstrates that the language pre-training helps the video instruction task with emergent language. Thus, the above ablation analysis shows that the joint training of emergent language and natural language can benefit each other.

4) \textit{Can emergent language provide more detailed and effective guidelines than video captions?}

In \model(+Cap, -EC), we replace the emergent language in \model with the language caption given by a MARN\cite{pei2019memory} as a data augmentation, which is a pre-trained video caption model. As shown in Table \ref{tab:ablation-study_whole}, \model shows an obvious advantage to \model(+Cap, -NL), which demonstrates that emergent language provides more detailed and effective guidelines than video captions.

5) \textit{What is the relationship between model size, control performance, and the correlation between emergent language and natural language?}

We first tested the overall average success rate of \model with different model sizes (listed in Table \ref{tab:model size}) in the downstream tasks (Franka Kitchen and MetaWorld). As shown in Figure \ref{fig:ablation_model_meta} and Figure \ref{fig:ablation_model_kitchen},  the larger the size of the model, the more expressive it is and thus the better its performance. 
Then, we perform a linear correlation analysis on the word embeddings of the emergent language and the word embeddings of the matched natural language.
We use the square of the coefficient of determination, denoted $R^{2}$ as the measure of predicting the performance of linear regression, which reflects the proportion of the variation in the dependent variable that is predictable from the independent variable. Consider a linear regression model $f(\cdot)$ which predicts $y$ with the input $x$, the coefficient of determination $R$ is defined as
\begin{equation}
R^2 = 1-\frac{SS_{res}}{SS_{tot}} = 1-\frac{\sum _{i}(y_{i}-f(x_{i}))^{2}}{\sum _{i}(y_{i}-{\bar {y}})^{2}}
\end{equation}
We evaluat the $R^{2}$ values of two linear regressions. One predicts natural language word embeddings from emergent language, and the other predicts emergent language word embeddings from natural language.
As shown in Figure \ref{fig:understand}, the $R^{2}$ of predicting natural language embeddings from the emergent language increase along with the model becoming larger, which corresponds to better performance. 
In contrast, the $R^{2}$ of predicting emergent language embeddings from the natural language is non-monotonous, along with the model size becoming larger. The reason is that emergent language captures more information that natural language does not contain when the model is large enough and has promising performance.
In short,   \model has the best performance in embodied control tasks when emergent language contains more information than natural language, the performance is median when emergent language and language information are nearly equal, and the performance is worst when emergent language contains less information than language information.
As shown in Table \ref{tab:Qualitative examples}, we also provide qualitative examples of emergent language and its corresponding natural language to show that emergent language can capture key concepts from video. We find that the tokens ``108" and ``500" correspond to ``drawer" in natural language and token ``701" corresponds to ``stapler." Except for nouns, emergent language can also get the concept of verbs, for example, token ``538"  corresponds to the verb ``reach" in natural language. The current manual check analysis of emergent language is just a preliminary attempt, and more understanding can be promising for future work.

\section{Conclusions and Discussions}
This paper aims to build the link between perceptual grounding and symbolic concepts by emergent communication (EC) language for embodied control. To this end, we develop a novel \textbf{E}mbodied \textbf{C}ontrol framework with the help of \textbf{E}mergent \textbf{C}ommunication language (\model), which pre-trains a language model via masked trajectory complement pretext task condition on emergent language or natural language. The pre-trained language model is utilized to extract embodied representation from instructions and observations and is used as a frozen module for downstream data-efficient policy learning.
Extensive experiments show that \textbf{\model} outperforms existing methods in Metaworld and Franka kitchen benchmarks. 
We believe that \model will serve as a solid step toward the general decision-making model.
\textbf{Limitations and future works:} more understanding about emergent language, training \model on a larger and more diverse dataset, and applications on more practical setup can be promising directions of future works.

\textbf{Acknowledgement.} Ping Luo is partially supported by the National Key R\&D Program of China No.2022ZD0161000 and the General Research Fund of HK No.17200622. Shunyu Yao was supported in part by the National Science Foundation under Grant No. 2107048. 

%

{\small
\bibliographystyle{ieee_fullname}
\bibliography{egbib}
}

\end{document}